\documentclass[letterpaper, 10 pt, conference]{ieeeconf}  

\IEEEoverridecommandlockouts                              

\usepackage{graphics} 
\usepackage{epsfig} 
\usepackage{mathptmx} 
\usepackage{times} 
\usepackage{amsmath} 
\usepackage{amssymb}  
\usepackage[noadjust]{cite}

\usepackage{booktabs}
\usepackage{enumerate}

\usepackage{graphicx, amsmath, amssymb, caption, subcaption, multirow, overpic, textpos}
\usepackage[table]{xcolor}
\usepackage{bbm}

\renewcommand{\paragraph}[1]{\vspace{1.25mm}\noindent\textbf{#1}}

\newcolumntype{x}[1]{>{\centering\arraybackslash}p{#1pt}}
\newcolumntype{y}[1]{>{\raggedright\arraybackslash}p{#1pt}}
\newcolumntype{z}[1]{>{\raggedleft\arraybackslash}p{#1pt}}

\newcommand{\app}{\raise.17ex\hbox{$\scriptstyle\sim$}}

\definecolor{deemph}{gray}{0.6}

\definecolor{baselinecolor}{gray}{.9}
\newcommand{\baseline}[1]{\cellcolor{baselinecolor}{#1}}

\definecolor{red}{HTML}{e74c3c}
\definecolor{green}{HTML}{2ecc71}

\usepackage{amsmath,bm}
\usepackage{graphicx}
\usepackage{dcolumn}
\usepackage[english]{babel}
\usepackage{subcaption,soul}
\usepackage{caption}
\usepackage[pagebackref=false,breaklinks=true,letterpaper=true,colorlinks=true,linkcolor = black, urlcolor = black,citecolor=black,bookmarks=true]{hyperref}
\usepackage{cleveref}
\usepackage{url}
\usepackage{multirow}
\usepackage{booktabs,comment,balance}
\usepackage{amsfonts}
\usepackage{float}
\usepackage{rotating}

\title{\LARGE \bf
Text Detection \& Recognition in the Wild for Robot Localization
}

\author{Zobeir Raisi$^{1}$ and John Zelek$^{1}$
\thanks{$^{1}$ Vision and Image Processing Lab, Systems Design Engineering Department,
        University of Waterloo, ON, Waterloo, N2L 3G1, Canada
        {\tt\small \{zraisi,jzelek\}@uwaterloo.ca}}%
}

\begin{document}

\maketitle
\thispagestyle{plain}
\pagestyle{plain}

\begin{abstract}


Signage is everywhere and a robot should be able to take advantage of signs to help it localize (including Visual Place Recognition (VPR)) and map.  Robust text detection \& recognition in the wild is challenging due such factors as pose, irregular text, illumination and occlusion.
We propose an end-to-end scene text spotting model that simultaneously outputs the text string and bounding boxes. This model is more suitable for VPR.
Our central contribution  is introducing utilizing an end-to-end scene text spotting framework to adequately capture the irregular and occluded text regions in different challenging places.
To evaluate our proposed architecture's performance for VPR, we conducted several experiments on the challenging Self-Collected Text Place (SCTP) benchmark dataset.
The initial experimental results show that the proposed method outperforms the SOTA methods in terms of precision and recall when tested on this benchmark.

\end{abstract}

\section{Introduction}
\label{sec:intro}

We live in a visual world, signage is everywhere.  Whether it is a street sign, a billboard, a house or room number or labels such as a license plate or a person's name; signage provides us useful information in terms of location and identity.  There have been many classifiers developed that are able to identify street signs or license plates with highly constrained priors on the method that do not allow their extension to general text in the wild detection and recognition.  However, to take advantage of all the signage available, we need to be able to detect signage (i.e., text) anywhere (i.e., in the wild).  OCR is a well solved problem for text detection and recognition in highly constrained environments however detecting and recognizing text anywhere is a challenging problem.

Signage can help a robot localize or map an environment.  Typically for SLAM processes, direct (i.e., pixel) or indirect features are used.  Signage can provide a coarse localization globally when the signage indicates an address or location.  Also the letters and numbers in a sign and their perspective can be used to determine relative pose if we assume the signage is on a planar surface or even just vertical wrt the ground plane. Visual Place Recognition (VPR) \cite{vpr1,vpr2,vpr3,vpr4,textplace_2019} aims to aid a vision guided system to localize wrt a previously visited place.  VPR has uses in loop closure detection for visual SLAM and localization in general.  Challenges in VPR include appearance variation due to perceptual aliasing, illumination, viewpoint changes, pose, weather, seasons to name a few.  Most techniques are focused on features (i.e., indirect) \cite{laguna2022key}  and sets of these features (e.g., BOW Bag of Words) methods.

Text spotting in the wild images is also called end-to-end scene text detection and recognition \cite{abcnetV2_2021,Raisi_2021_CVPR}.
Simultaneous text detection and recognition go hand in hand. In scene text detection, the goal is to localize words in the image, and for scene text recognition, the aim is to convert the patch of cropped word images into a sequence of characters. Like scene text detection and recognition tasks, scene text spotting also encounters different challenging problems, including irregular text, illumination variations, low-resolution text, occlusion, \textit{etc} \cite{zobeir_2020}.

Previous methods in scene text detection and recognition have utilized convolutional neural network (CNN) as a feature extractor \cite{fasterrcnn_ren2015,liu2016ssd,YOLO_2016,Maskrcnn_He2017} and Recurrent Neural Networks (RNN) \cite{RNN_1986,LSTM_1997,baek2019STR} for capturing sequential dependency. 
Despite achieving promising performances on various challenging benchmark datasets \cite{lucas2003icdar,wang2010word,shahab2011icdar,Mishra_12_IIIT,yao2012detecting,svtp_2013,karatzas2013icdar,cut80_2014,lin2014microsoft,karatzas2015icdar,icdar2017,ch2017total,CTW_1500_yuliang2017,gupta2016synthetic,jaderberg2014synthetic}, it has been shown that there are two main challenges for detecting or recognizing text in the wild images that has been studied in the past years. (1) Irregular text refers to the text with arbitrary shapes that usually have severe orientation and curvature, and (2) occlusion, which makes poor performance on the existing methods \cite{shi2018aster,baek2019craft,pmtd_liu_2019,baek2019STR} due to their reliance on the visibility of the target characters in the given images.
Furthermore, CNN's have two significant drawbacks: (1) they have problems in capturing long-range dependencies (e.g., arbitrary relations between pixels in spatial domains) due to their fixed-size window operation \cite{deformableDETR_2020}, (2) they suffer from dynamical adoption to the changes to the inputs because the convolution filter weights are tuned to a specific training distribution \cite{Transformer_survey2_2020}.

Recent end-to-end scene text spotting methods \cite{raisi_cvis_2022,TTS_2022,TESTR_2022,Raisi_2022_CRV} utilized transformers \cite{attention_vaswani2017} in their architecture and achieved superior performance in many benchmarks \cite{ch2017total,CTW_1500_yuliang2017}.
Transformers \cite{attention_vaswani2017}, and their variations \cite{performer_2020,detr_2020,deformableDETR_2020}, are a new deep learning architecture that mitigates the issues mentioned above for CNNs;
{Unlike Recurrent Neural Networks (RNNs),
transformers are models that learn how to encode and decode data by looking not only backward but also forward to extract relevant information from a whole sequence allowing conducting complex tasks such as machine translation} \cite{attention_vaswani2017}, speech recognition \cite{chan2020imputer}, and recently in computer vision \cite{detr_2020,dosovitskiy2020image,Transformer_survey2_2020}.
The attention mechanism allows the transformers to reason more effectively and focus on the relevant parts of the input data (e.g., a word in a sentence for machine translation and a character of a word in a text image for detection and recognition) as needed.

Visual place recognition (VPR) 
\cite{textplace_2019}  aims to recognize the previously visited places using visual information with resilience to perceptual aliasing, illumination, and viewpoint changes. Text that appears in the wild images, such as street signs, billboards, and shop signage, usually carries extensive discriminative information. VPR task can take advantage of these scene texts with high-level information for previously visited place recognition.

This paper leverages a pre-trained end-to-end transformer-based text spotting framework for the VPR task. Unlike \cite{textplace_2019} that used two separate modules of detection and recognition for extracting the text regions, our model can directly read the text instances from the given frame in an end-to-end manner. Furthermore, by equipping a masked autoencoder (MAE) \cite{mae_2021} as a backbone, our proposed model is more robust in capturing occluded text instance regions, which makes it more suitable for visual place recognition. 
Our main contributions are as follows: (1) We utilize an end-to-end transformer-based scene text spotting pipeline for the VPR application for the first time. Our model can handle arbitrary shapes text with polygon bounding boxes and output word instances simultaneously. (2) We provide a quantitative and qualitative comparison of our method with state-of-the-art (SOTA) techniques.


\begin{figure*}
\vspace{-15pt}
    \centering
    \includegraphics[width=0.92\linewidth]{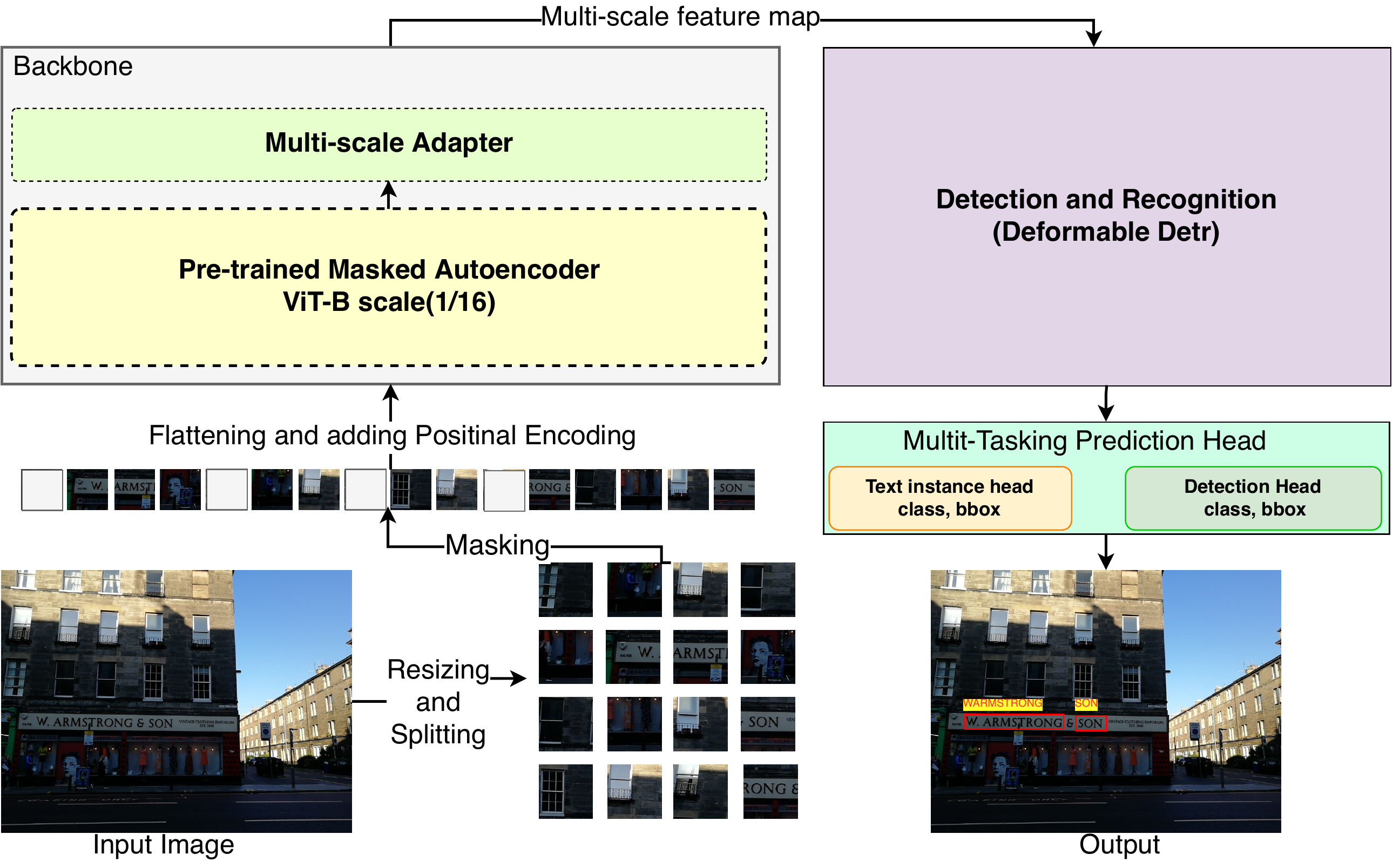}
    \caption{Block diagram of the proposed scene text spotting architecture using transformer for VPR \cite{Raisi_2022_CRV}. Unlike the step-wise pipeline in \cite{textplace_2019}, our model output the bounding box coordinates and the word instances in an end-to-end manner for the VPR task (See section \ref{sec:methodlogy} for more details). Best viewed when zoomed.}
    \label{fig:proposed_arch}
    \vspace{-15pt}
\end{figure*}

\section{Related work}
\label{sec:related}

\subsection{End-to-end Scene Text Detection and Recognition} 
\label{sec:textDetection}

Scene text spotting aims to detect and recognize text instances from a given image end-to-end 
\cite{TowardsET,fasterrcnn_ren2015,liu2018fots,lyu2018mask,textdragon_2019,qin19towards,spotterV3,abcnet_2020}.  Similar to different computer vision tasks, deep learning techniques using
CNN/RNN-based methods and transformer-based methods are dominant frameworks in scene text spotting. 
%

Early methods \cite{TowardsET} in scene text spotting have mainly utilized a deep-learning convolutional neural network (CNN) as a feature extractor \cite{ResNet_He2015L} and Recurrent Neural Networks (RNN) \cite{RNN_1986,LSTM_1997,baek2019STR} to read horizontal scene text.
For example, Li \textit{et al.} \cite{TowardsET} combined the detection and recognition framework to present the first text spotting method by using a shared CNN backbone encoder, following by RoIPooling \cite{fasterrcnn_ren2015} as detection. Then the resulted features are fed into RNN recognition to output the final word instances for a given input image. FOTS \cite{liu2018fots} utilized an anchor-free CNN-based object detection framework that improved both the training and inference time. It also used a RoIRotate module for reading rotated text instances.

Since text in the wild images appear in arbitrary shapes including multi-oriented and curved, several methods \cite{lyu2018mask,textdragon_2019,qin19towards,spotterV3,abcnet_2020} targeted reading these type of text instances. These methods usually used a CNN-based segmentation network with multiple post-processing stages to output polygon box coordinates for the final irregular texts.
For instance, in \cite{qin19towards}, a RoiMask is used to connect both the detection and recognition module for capturing arbitrary shaped text. Liu \cite{abcnet_2020} leveraged a Bezier curve representation for the detection part, followed by a Bezier Align module to rectify the curved text instances into a regular text before feeding it to the attention-based recognition part. 
Some methods \cite{crafts_2020,raisi_cvis_2022} targeted spotting individual characters and merging them to output the final arbitrary shape text instance. 

Recently with the advancement of transformers \cite{attention_vaswani2017} in computer vision fields \cite{transformer_survey_2020,survey_visualTransforumer,Transformer_survey2_2020}, several  SOTA scene text spotting methods \cite{CVIS_21,SATRN_2020,vitstr_2021,Raisi_2021_CRV,ABINet_2021,zobeir_icpr} proposed to take the benefit of transformer-based pipelines in their framework. These methods achieved superior performance in both regular and irregular benchmark datasets. 
For example, Kittenplon \textit{et al.} \cite{TTS_2022} utilized a transformer-based detector, Deformable-DETR's \cite{deformableDETR_2020}, as its primary framework by proposing a  multi-task prediction head that  can output word instances and box coordinates of an arbitrary shape text.
\cite{TESTR_2022} used  transformers as the main block for an end-to-end text spotting framework for text detection and recognition in wild images. These methods removed the dependency of region-of-interest operations and post-processing stages in their framework. Thus, they  can output both Bezier curve and polygon representations and achieve superior benchmark performance.
Very recently, Raisi \textit{et al.} \cite{Raisi_2022_CRV} proposed an end-to-end framework for scene text spotting that is also capable of improving the recognition performance for an adverse situation like occlusion. This method utilized an MAE in their pipeline equipped with a powerful detector, namely deformable-DETR \cite{deformableDETR_2020}, to capture the arbitrary shape of occluded text instance in the wild images. In this work, we use a pre-trained model of \cite{Raisi_2022_CRV} for the VPR task.

\vspace{-3pt}
\section{Methodology}
\label{sec:methodlogy}

For a complete text reading, simultaneous text detection and recognition are required. 
Unlike step-wise detection and recognition, the end-to-end framework will improve the overall speed by eliminating multiple processing steps. Furthermore, an end-to-end transformer is expected to offer higher accuracy compared to previous end-to-end CNN-based approaches \cite{liu2018fots,abcnet_2020}.

\subsection{Scene Text Spotting Architecture}
The overall framework of our proposed method is shown in Figure \ref{fig:proposed_arch}.

\vspace{-3pt}
\paragraph{Backbone:}
Inspired \cite{mae_bench2021}, the proposed model uses a pre-trained models of the Vision Transformer architecture (ViT) \cite{dosovitskiy2020image} as the backbone. The input image is first split into a non-overlapping sequence of patches.
After masking a large set of the input patches ($\sim 75\%$) and adding the $1$D position embedding, these patches are passed into transformer blocks containing several multi-head self-attention and feed-forward modules.
However, the final output of the transformer encoder backbone is single-scale due to the columnar structure of ViT, which makes them inadequate for detecting multi-scale text instances. To address this, we utilize a multi-scale adapter module \cite{mae_bench2021}. It is worth mentioning that we use a pre-trained MAE \cite{mae_2021} (ViT-Base/16) as the backbone for feature extraction. This backbone was further fine-tuned on 36 classes of alphanumeric characters.

\vspace{-3pt}
\paragraph{Multi-scale adapter:} Inspired from \cite{mae_bench2021,Raisi_2022_CRV}, we adapt the single-scale ViT into the multi-scale FPN for capturing different resolution of text regions. The multi-scale feature map module utilizes the idea of up-sampling or down-sampling into the intermediate single-scale ViT’s feature map with columnar structure \cite{mae_bench2021}. The resulted multi-scale feature maps are then fed into a modified detector \cite{deformableDETR_2020} for detecting and recognizing word instances.

\vspace{-3pt}
\paragraph{Text predictor:}
After feature extraction and multi-scaling, the resulted feature maps are fed to the text of the final module to detect and recognize the text instance of a given image. As shown in Figure \ref{fig:proposed_arch}, the proposed text predictor leverages a modified Deformable-DETR \cite{deformableDETR_2020} with multi-task prediction head.
During training, the encoder's multi-head self-attention detector learns how to separate individual character and word instances in the scene image by performing global computations. The decoder typically learns how to attend to a different part of characters in words by using different learn-able vectors (so-called object queries). After training, the multi-task head (last layer of the decoder) can directly predict both absolute bounding box coordinates and sequence of characters, eliminating the use of any hand-designed components and post-processing like anchor design and non-max suppression.
After reading the text instances from the scene images, we implement the same text filtering criteria as introduced in \cite{textplace_2019} for comparing the query and inference frames.

\begin{figure*}
    \centering
    \includegraphics[width=\linewidth]{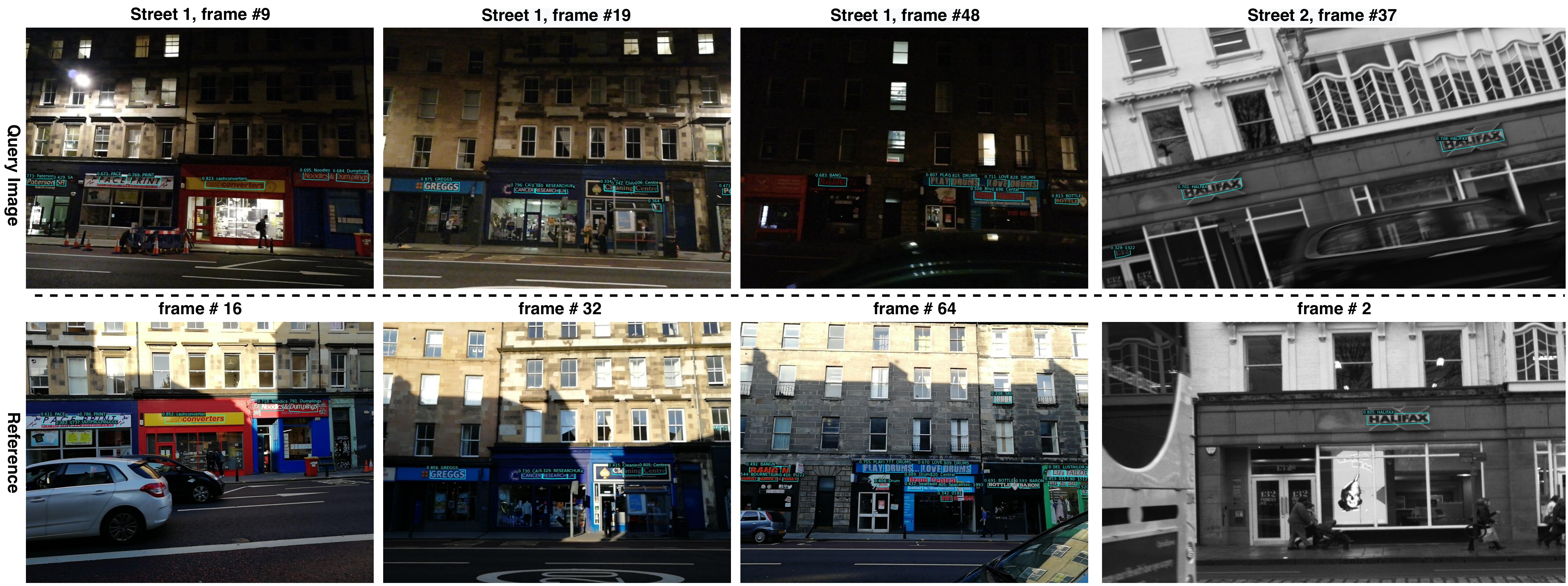}
    \caption{Query image and matching reference examples of \cite{textplace_2019} dataset. Our proposed model detects and recognizes the most challenging text instances required to match the query (top column) and reference (bottom column) frames. Best viewed in color when zoomed.}
    \label{fig:qual1}
\end{figure*}

\begin{figure}
\vspace{-15pt}
    \centering
    \includegraphics[width=0.99\linewidth]{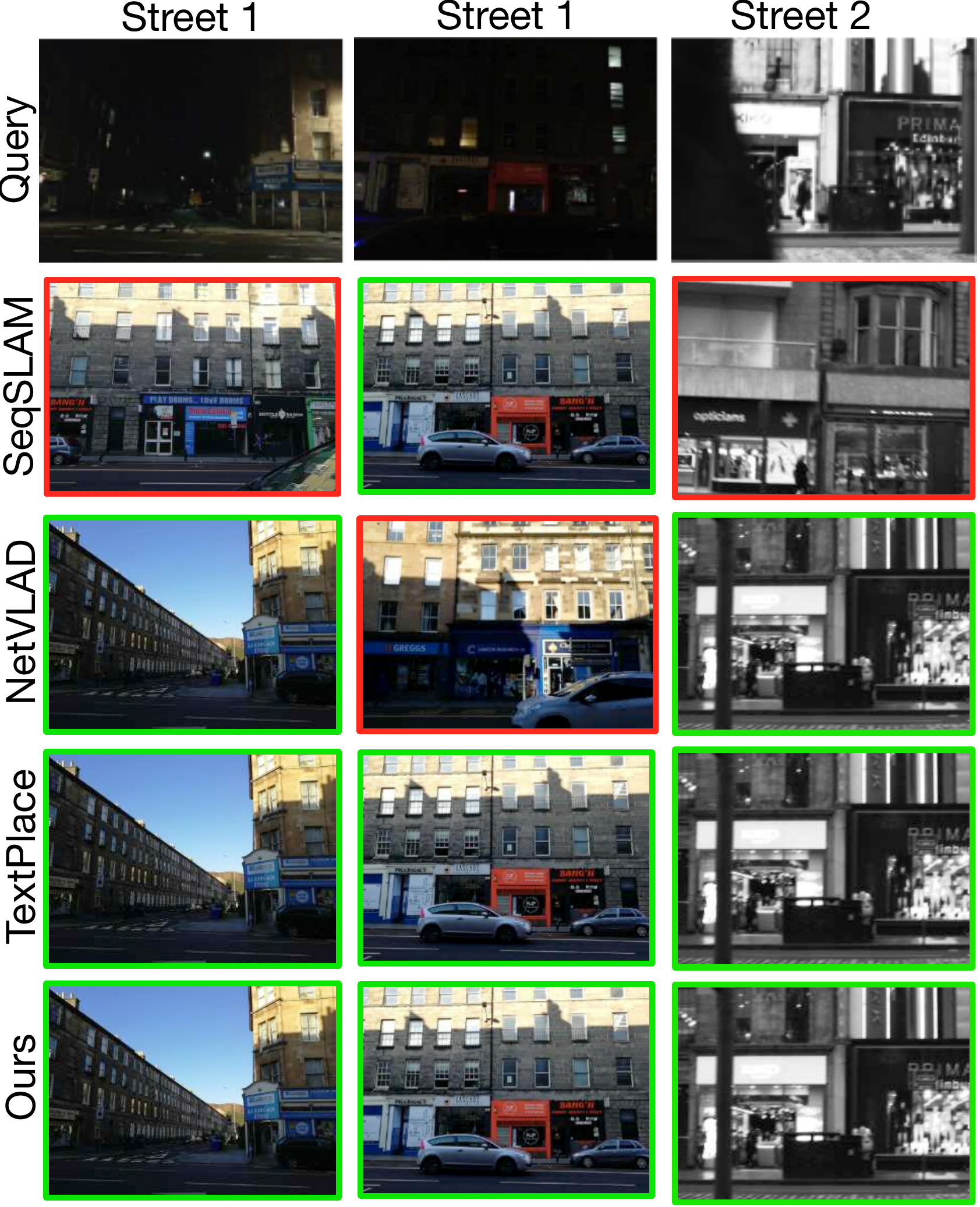}
    \caption{Qualitative comparison of the proposed model with SOTA methods \cite{seqslam_2012,netvlad_2016,textplace_2019} on the SCTP dataset. The correct and incorrect results are bounded with green and red colors.}
    \label{fig:qual2}
    \vspace{-15pt}
\end{figure}

\section{Experimental Result}
\label{sec:experiment}

\subsection{Datasets and Evaluation Metric}
\vspace{-3pt}
The
Self-Collected TextPlace (SCTP) Dataset \cite{textplace_2019}
is designed explicitly for visual place recognition tasks in urban places. The images of this dataset are captured using a side-looking mobile phone camera. These images include three pairs of map and query sequences in outdoor streets and an indoor shopping mall and 
contain significant challenging scenarios, including high dynamics, random occlusions, severe illumination changes, irregular text instances, and viewpoint changes.
To compare the performance of our proposed model with SOTA methods \cite{textplace_2019,TodayGAN_2019,netvlad_2016}, we use the precision recall evaluation measurement described in \cite{icdar2019,textplace_2019}.

\subsection{Quantitative Comparison}

The quantitative results of our proposed model with several SOTA methods \cite{FAB-MAP_2008,seqslam_2012,netvlad_2016,TodayGAN_2019,textplace_2019} on SCTP datasets \cite{textplace_2019} are shown in Table \ref{tab:quant}. Our  model achieved the best performance in terms of recall for this dataset, which contains significant challenging like irregular and partially occluded text instances. This performance confirms the effectiveness of our proposed method for VPR.

\begin{table}[]
    \centering
    \caption{Precision-Recall comparison of our proposed model with SOTA methods including TextPlace \cite{textplace_2019}, ToDayGAN \cite{TodayGAN_2019}, NetVLAD \cite{netvlad_2016}, SeqSLAM \cite{seqslam_2012}, and FAB-MAP \cite{FAB-MAP_2008} using SCTP \cite{textplace_2019} dataset. The best performance is highlighted in \bf{bold}.}
    \resizebox{\linewidth}{!}{%
    \begin{tabular}{l|ccccc} \hline
    Model & \multicolumn{5}{c}{Recall}\\ \cline{2-6}
    &0.2 &0.4& 0.6& 0.8& 0.9\\ \hline
    \baseline{\textbf{Our model}}        &\baseline{1} & \baseline{1} &\baseline{1} &\baseline{\textbf{0.97}} &\baseline{\textbf{0.93}}  \\
    TextPlace   &1 & 1 &1 &0.96 &0.91 \\
    NetVLAD-10  &1 & 1 &1 &0.95 &\textbf{0.93} \\
    NetVLAD-20  &1 & 1 &1 &0.91 &0.87 \\
    NetVLAD-30  &1 & 1 &0.97 &0.85& 0.83 \\
    ToDayGAN-10 &0.50 &0.55 &0.58 &0.57 &0.56 \\
    ToDayGAN-20 &0.40 &0.40 &0.40 &0.38 &0.38 \\
    ToDayGAN-30 &0.26 &0.24 &0.24 &0.25 &0.24 \\
    FAB-MAP-10  &0.79 &0.69 &0.67 &0.65 &0.63 \\
    FAB-MAP-20  &0.76 &0.69 &0.67 &0.63 &0.60 \\
    FAB-MAP-30  &0.68 &0.67 &0.67 &0.62 &0.58 \\
    SeqSLAM     &0.30 &0.24 &0.18 &0.13 &0.13 \\ \hline
    \end{tabular}}
    \label{tab:quant}
\end{table}

\subsection{Qualitative Results}
Figure \ref{fig:qual1} illustrates qualitative results for the  \cite{textplace_2019} dataset. Our model successfully read challenging text instances of both query and reference frames. We also compare our model with some of SOTA techniques \cite{seqslam_2012,netvlad_2016,textplace_2019} in Figure \ref{fig:qual2}; As shown, our proposed text spotting model matches correctly the query frame with frame in inference.

We also experiment with measuring the inference speed of our model and comparing it with \cite{textplace_2019} in terms of Frames Per Second (FPS). To that effect and for a fair comparison, we use an RTX 3080Ti GPU that has a similar memory used in \cite{textplace_2019} and presented in \cite{liao2018textboxes++}; Our model outperformed the TextPlace method by a large margin achieving $\sim 11$ FPS in compare to $2.3$ FPS in \cite{textplace_2019}.

It is worth mentioning that our proposed model performs poorly on extremely low-resolution and blurry text instances. However, these challenges are still open in many SOTA text detection and recognition in the wild methods.

\section{CONCLUSIONS}
\vspace{-3pt}
We presented an end-to-end scene text spotting model for the visual place recognition task. The proposed model has leveraged a robust and SOTA backbone of pre-trained MAE and a modified multi-task transformer detector.
%
Our experimental results have shown that the proposed model outperforms SOTA models in VPR, which confirms the robustness of our end-to-end scene text detection and recognition model. 
Other applications besides VPR include different facets of localization and mapping.  Being able to detect and recognize text allows the potential to leverage semantics and the features related to the detected text to better localize and map as opposed to just using indirect features \cite{laguna2022key}.



\section*{ACKNOWLEDGMENT}
We would like to thank the Ontario Centres of Excellence (OCE),
the Natural Sciences and Engineering Research Council of Canada
(NSERC), and ATS Automation Tooling Systems Inc., Cambridge,
ON, Canada for supporting this research work.
We provide extra experiments in Appendix (See \S \ref{sec:appendix}).

\addtolength{\textheight}{12cm}   



\addtolength{\textheight}{-12 cm}
{
\bibliographystyle{IEEEtranS}
\bibliography{IEEEexample}

\begin{thebibliography}{10}
\providecommand{\url}[1]{#1}
\csname url@rmstyle\endcsname
\providecommand{\newblock}{\relax}
\providecommand{\bibinfo}[2]{#2}
\providecommand\BIBentrySTDinterwordspacing{\spaceskip=0pt\relax}
\providecommand\BIBentryALTinterwordstretchfactor{4}
\providecommand\BIBentryALTinterwordspacing{\spaceskip=\fontdimen2\font plus
\BIBentryALTinterwordstretchfactor\fontdimen3\font minus
  \fontdimen4\font\relax}
\providecommand\BIBforeignlanguage[2]{{%
\expandafter\ifx\csname l@#1\endcsname\relax
\typeout{** WARNING: IEEEtran.bst: No hyphenation pattern has been}%
\typeout{** loaded for the language `#1'. Using the pattern for}%
\typeout{** the default language instead.}%
\else
\language=\csname l@#1\endcsname
\fi
#2}}

\bibitem{TodayGAN_2019}
A.~Anoosheh, T.~Sattler, R.~Timofte, M.~Pollefeys, and L.~Van~Gool,
  ``Night-to-day image translation for retrieval-based localization,'' in
  \emph{2019 International Conference on Robotics and Automation (ICRA)}, 2019,
  pp. 5958--5964.

\bibitem{netvlad_2016}
R.~Arandjelovic, P.~Gronat, A.~Torii, T.~Pajdla, and J.~Sivic, ``Netvlad: Cnn
  architecture for weakly supervised place recognition,'' in \emph{Proc.
  IEEE/CVF Intl. Conf. on Comp. Vision}, 2016, pp. 5297--5307.

\bibitem{vitstr_2021}
R.~Atienza, ``Vision transformer for fast and efficient scene text
  recognition,'' in \emph{Document Analysis and Recognition -- ICDAR
  2021}.\hskip 1em plus 0.5em minus 0.4em\relax Springer International
  Publishing, 2021, pp. 319--334.

\bibitem{baek2019STR}
J.~Baek, G.~Kim, J.~Lee, S.~Park, D.~Han, S.~Yun, S.~J. Oh, and H.~Lee, ``What
  is wrong with scene text recognition model comparisons? dataset and model
  analysis,'' in \emph{Proc. Int. Conf. on Comp. Vision (ICCV)}, 2019.

\bibitem{baek2019craft}
Y.~Baek, B.~Lee, D.~Han, S.~Yun, and H.~Lee, ``Character region awareness for
  text detection,'' in \emph{Proc. IEEE Conf. on Comp. Vision and Pattern
  Recognit.}, 2019.

\bibitem{crafts_2020}
Y.~Baek, S.~Shin, J.~Baek, S.~Park, J.~Lee, D.~Nam, and H.~Lee, ``Character
  region attention for text spotting,'' \emph{ArXiv}, vol. abs/2007.09629,
  2020.

\bibitem{detr_2020}
N.~Carion, F.~Massa, G.~Synnaeve, N.~Usunier, A.~Kirillov, and S.~Zagoruyko,
  ``End-to-end object detection with transformers,'' \emph{arXiv preprint
  arXiv:2005.12872}, 2020.

\bibitem{chan2020imputer}
W.~Chan, C.~Saharia, G.~Hinton, M.~Norouzi, and N.~Jaitly, ``Imputer: Sequence
  modelling via imputation and dynamic programming,'' \emph{arXiv preprint
  arXiv:2002.08926}, 2020.

\bibitem{ch2017total}
C.~K. Ch'ng and C.~S. Chan, ``Total-text: A comprehensive dataset for scene
  text detection and recognition,'' in \emph{Proc. IAPR Int. Conf. on Document
  Anal. and Recognit. (ICDAR)}, vol.~1, 2017, pp. 935--942.

\bibitem{performer_2020}
K.~Choromanski, V.~Likhosherstov, D.~Dohan, X.~Song, A.~Gane, T.~Sarlos,
  P.~Hawkins, J.~Davis, A.~Mohiuddin, L.~Kaiser, \emph{et~al.}, ``Rethinking
  attention with performers,'' \emph{arXiv preprint arXiv:2009.14794}, 2020.

\bibitem{FAB-MAP_2008}
M.~Cummins and P.~Newman, ``Fab-map: Probabilistic localization and mapping in
  the space of appearance,'' \emph{The International Journal of Robotics
  Research}, vol.~27, no.~6, pp. 647--665, 2008.

\bibitem{dosovitskiy2020image}
A.~Dosovitskiy, L.~Beyer, A.~Kolesnikov, D.~Weissenborn, X.~Zhai,
  T.~Unterthiner, M.~Dehghani, M.~Minderer, G.~Heigold, S.~Gelly,
  \emph{et~al.}, ``An image is worth 16x16 words: Transformers for image
  recognition at scale,'' \emph{arXiv preprint arXiv:2010.11929}, 2020.

\bibitem{ABINet_2021}
S.~Fang, H.~Xie, Y.~Wang, Z.~Mao, and Y.~Zhang, ``Read like humans: Autonomous,
  bidirectional and iterative language modeling for scene text recognition,''
  in \emph{Proc. IEEE/CVF Conf. on Comput. Vision and Pattern Recognit.}, 2021,
  pp. 7098--7107.

\bibitem{textdragon_2019}
W.~Feng, W.~He, F.~Yin, X.-Y. Zhang, and C.-L. Liu, ``Textdragon: An end-to-end
  framework for arbitrary shaped text spotting,'' in \emph{Proc. IEEE/CVF
  Confon. Comput. Vision and Pattern Recognit.}, 2019, pp. 9076--9085.

\bibitem{vpr1}
S.~Garg, T.~Fischer, and M.~Milford, ``Where is your place, visual place
  recognition?'' \emph{arXiv preprint arXiv:2103.06443}, 2021.

\bibitem{gupta2016synthetic}
A.~Gupta, A.~Vedaldi, and A.~Zisserman, ``Synthetic data for text localisation
  in natural images,'' in \emph{Proc. IEEE Conf. on Comp. Vision and Pattern
  Recognit.}, 2016, pp. 2315--2324.

\bibitem{survey_visualTransforumer}
K.~Han, Y.~Wang, H.~Chen, X.~Chen, J.~Guo, Z.~Liu, Y.~Tang, A.~Xiao, C.~Xu,
  Y.~Xu, \emph{et~al.}, ``A survey on visual transformer,'' \emph{arXiv
  preprint arXiv:2012.12556}, 2020.

\bibitem{mae_2021}
K.~He, X.~Chen, S.~Xie, Y.~Li, P.~Doll{\'a}r, and R.~Girshick, ``Masked
  autoencoders are scalable vision learners,'' \emph{arXiv preprint
  arXiv:2111.06377}, 2021.

\bibitem{Maskrcnn_He2017}
K.~He, G.~Gkioxari, P.~Doll{\'a}r, and R.~Girshick, ``Mask \textup{R-CNN},'' in
  \emph{Proc. IEEE Int. Conf. on Comp. Vision}, 2017, pp. 2961--2969.

\bibitem{ResNet_He2015L}
K.~He, X.~Zhang, S.~Ren, and J.~Sun, ``Deep residual learning for image
  recognition,'' \emph{Proc. IEEE Conf. on Comp. Vision and Pattern Recognit.
  (CVPR)}, pp. 770--778, 2015.

\bibitem{LSTM_1997}
S.~Hochreiter and J.~Schmidhuber, ``Long short-term memory,'' \emph{Neural
  computation}, vol.~9, no.~8, pp. 1735--1780, 1997.

\bibitem{textplace_2019}
Z.~Hong, Y.~Petillot, D.~Lane, Y.~Miao, and S.~Wang, ``Textplace: Visual place
  recognition and topological localization through reading scene texts,'' in
  \emph{Proc. IEEE/CVF Intl. Conf. on Comp. Vision}, 2019, pp. 2861--2870.

\bibitem{icdar2017}
M.~Iwamura, N.~Morimoto, K.~Tainaka, D.~Bazazian, L.~Gomez, and D.~Karatzas,
  ``\textup{ICDAR2017} robust reading challenge on omnidirectional video,'' in
  \emph{Proc. IAPR Int. Conf. on Document Anal. and Recognition (ICDAR)},
  vol.~1, 2017, pp. 1448--1453.

\bibitem{jaderberg2014synthetic}
M.~Jaderberg, K.~Simonyan, A.~Vedaldi, and A.~Zisserman, ``Synthetic data and
  artificial neural networks for natural scene text recognition,'' \emph{arXiv
  preprint arXiv:1406.2227}, 2014.

\bibitem{karatzas2015icdar}
D.~Karatzas, L.~Gomez-Bigorda, A.~Nicolaou, S.~Ghosh, A.~Bagdanov, M.~Iwamura,
  J.~Matas, L.~Neumann, V.~R. Chandrasekhar, S.~Lu, \emph{et~al.},
  ``\textup{ICDAR} 2015 competition on robust reading,'' in \emph{Proc. Int.
  Conf. on Document Anal. and Recognition (ICDAR)}, 2015, pp. 1156--1160.

\bibitem{karatzas2013icdar}
D.~Karatzas, F.~Shafait, S.~Uchida, M.~Iwamura, L.~G. i~Bigorda, S.~R. Mestre,
  J.~Mas, D.~F. Mota, J.~A. Almazan, and L.~P. De~Las~Heras, ``\textup{ICDAR}
  2013 robust reading competition,'' in \emph{Proc. Int. Conf. on Document
  Anal. and Recognition}, 2013, pp. 1484--1493.

\bibitem{Transformer_survey2_2020}
S.~Khan, M.~Naseer, M.~Hayat, S.~W. Zamir, F.~S. Khan, and M.~Shah,
  ``Transformers in vision: A survey,'' \emph{arXiv preprint arXiv:2101.01169},
  2021.

\bibitem{TTS_2022}
Y.~Kittenplon, I.~Lavi, S.~Fogel, Y.~Bar, R.~Manmatha, and P.~Perona, ``Towards
  weakly-supervised text spotting using a multi-task transformer,'' \emph{arXiv
  preprint arXiv:2202.05508}, 2022.

\bibitem{laguna2022key}
A.~B. Laguna and K.~Mikolajczyk, ``Key. net: Keypoint detection by handcrafted
  and learned cnn filters revisited,'' \emph{IEEE Transactions on Pattern
  Analysis and Machine Intelligence}, 2022.

\bibitem{SATRN_2020}
J.~Lee, S.~Park, J.~Baek, S.~Joon~Oh, S.~Kim, and H.~Lee, ``On recognizing
  texts of arbitrary shapes with \textup{2D} self-attention,'' in \emph{IEEE
  CVPR}, 2020, pp. 546--547.

\bibitem{TowardsET}
H.~Li, P.~Wang, and C.~Shen, ``Towards end-to-end text spotting with
  convolutional recurrent neural networks,'' \emph{2017 IEEE International
  Conference on Computer Vision (ICCV)}, pp. 5248--5256, 2017.

\bibitem{mae_bench2021}
Y.~Li, S.~Xie, X.~Chen, P.~Dollar, K.~He, and R.~Girshick, ``Benchmarking
  detection transfer learning with vision transformers,'' \emph{arXiv preprint
  arXiv:2111.11429}, 2021.

\bibitem{spotterV3}
M.~Liao, G.~Pang, J.~Huang, T.~Hassner, and X.~Bai, ``Mask textspotter v3:
  Segmentation proposal network for robust scene text spotting,'' in
  \emph{Computer Vision--ECCV 2020: 16th European Conference, Glasgow, UK,
  August 23--28, 2020, Proceedings, Part XI 16}, 2020, pp. 706--722.

\bibitem{liao2018textboxes++}
M.~Liao, B.~Shi, and X.~Bai, ``Textboxes++: A single-shot oriented scene text
  detector,'' \emph{IEEE Trans. on Image process.}, vol.~27, no.~8, pp.
  3676--3690, 2018.

\bibitem{lin2014microsoft}
T.-Y. Lin, M.~Maire, S.~Belongie, J.~Hays, P.~Perona, D.~Ramanan,
  P.~Doll{\'a}r, and C.~L. Zitnick, ``Microsoft coco: Common objects in
  context,'' in \emph{Proc. Eur. Conf. on Comp. Vision}.\hskip 1em plus 0.5em
  minus 0.4em\relax Springer, 2014, pp. 740--755.

\bibitem{pmtd_liu_2019}
J.~Liu, X.~Liu, J.~Sheng, D.~Liang, X.~Li, and Q.~Liu, ``Pyramid mask text
  detector,'' \emph{CoRR}, vol. abs/1903.11800, 2019.

\bibitem{liu2016ssd}
W.~Liu, D.~Anguelov, D.~Erhan, C.~Szegedy, S.~Reed, C.-Y. Fu, and A.~C. Berg,
  ``\textup{SSD}: Single shot multibox detector,'' in \emph{Eur. Conf. on Comp.
  Vision}.\hskip 1em plus 0.5em minus 0.4em\relax Springer, 2016, pp. 21--37.

\bibitem{liu2018fots}
X.~Liu, D.~Liang, S.~Yan, D.~Chen, Y.~Qiao, and J.~Yan, ``\textup{FOTS: F}ast
  oriented text spotting with a unified network,'' in \emph{Proc. IEEE Conf. on
  Comp. Vision and Pattern Recognit.}, 2018, pp. 5676--5685.

\bibitem{abcnet_2020}
Y.~Liu, H.~Chen, C.~Shen, T.~He, L.~Jin, and L.~Wang, ``Abcnet: Real-time scene
  text spotting with adaptive bezier-curve network,'' in \emph{Proc. IEEE/CVF
  Conf. on Comput. Vision and Pattern Recognit.}, 2020, pp. 9809--9818.

\bibitem{abcnetV2_2021}
Y.~Liu, C.~Shen, L.~Jin, T.~He, P.~Chen, C.~Liu, and H.~Chen, ``Abcnet v2:
  Adaptive bezier-curve network for real-time end-to-end text spotting,''
  \emph{arXiv preprint arXiv:2105.03620}, 2021.

\bibitem{vpr3}
S.~Lowry, N.~S{\"u}nderhauf, P.~Newman, J.~J. Leonard, D.~Cox, P.~Corke, and
  M.~J. Milford, ``Visual place recognition: A survey,'' \emph{IEEE
  Transactions on Robotics}, vol.~32, no.~1, pp. 1--19, 2015.

\bibitem{lucas2003icdar}
S.~M. Lucas, A.~Panaretos, L.~Sosa, A.~Tang, S.~Wong, and R.~Young,
  ``\textup{ICDAR} 2003 robust reading competitions,'' in \emph{Seventh Int.
  Conf. on Document Anal. and Recognition, 2003. Proceedings.}, 2003, pp.
  682--687.

\bibitem{lyu2018mask}
P.~Lyu, M.~Liao, C.~Yao, W.~Wu, and X.~Bai, ``Mask textspotter: An end-to-end
  trainable neural network for spotting text with arbitrary shapes,'' in
  \emph{Proc. Eur. Conf. on Comp. Vision (ECCV)}, 2018, pp. 67--83.

\bibitem{vpr2}
C.~Masone and B.~Caputo, ``A survey on deep visual place recognition,''
  \emph{IEEE Access}, vol.~9, pp. 19\,516--19\,547, 2021.

\bibitem{seqslam_2012}
M.~J. Milford and G.~F. Wyeth, ``Seqslam: Visual route-based navigation for
  sunny summer days and stormy winter nights,'' in \emph{Proc. IEEE Intl. Conf.
  on Robotics and Automation}, 2012, pp. 1643--1649.

\bibitem{Mishra_12_IIIT}
A.~Mishra, K.~Alahari, and C.~V. Jawahar, ``Scene text recognition using higher
  order language priors,'' in \emph{BMVC}, 2012.

\bibitem{qin19towards}
S.~Qin, A.~Bissacco, M.~Raptis, Y.~Fujii, and Y.~Xiao, ``Towards unconstrained
  end-to-end text spotting,'' in \emph{Proc. IEEE/CVF Intl. Conf. on Computer
  Vision}, 2019, pp. 4704--4714.

\bibitem{svtp_2013}
T.~Quy~Phan, P.~Shivakumara, S.~Tian, and C.~Lim~Tan, ``Recognizing text with
  perspective distortion in natural scenes,'' in \emph{Proc. IEEE Intl. Conf.
  on Comput. Vision}, 2013, pp. 569--576.

\bibitem{CVIS_21}
Z.~Raisi, M.~Naiel, P.~Fieguth, S.~Wardell, and J.~Zelek, ``2d positional
  embedding-based transformer for scene text recognition,'' \emph{Journal of
  Computational Vision and Imaging Systems}, vol.~6, no.~1, p. 1–4, Jan.
  2021.

\bibitem{zobeir_2020}
Z.~Raisi, M.~A. Naiel, P.~Fieguth, S.~Wardell, and J.~Zelek, ``Text detection
  and recognition in the wild: A review,'' \emph{arXiv preprint
  arXiv:2006.04305}, 2020.

\bibitem{Raisi_2021_CRV}
Z.~Raisi, M.~A. Naiel, G.~Younes, S.~Wardell, and J.~Zelek, ``2lspe: 2d
  learnable sinusoidal positional encoding using transformer for scene text
  recognition,'' in \emph{Proc. Conf. on Robots and Vision (CRV)}, 2021, pp.
  119--126.

\bibitem{Raisi_2021_CVPR}
Z.~Raisi, M.~A. Naiel, G.~Younes, S.~Wardell, and J.~S. Zelek,
  ``Transformer-based text detection in the wild,'' in \emph{Proc. IEEE/CVF
  Conference on Computer Vision and Pattern Recognition (CVPR) Workshops}, June
  2021, pp. 3162--3171.

\bibitem{zobeir_icpr}
Z.~Raisi, G.~Younes, and J.~Zelek, ``Arbitrary shape text detection using
  transformers,'' in \emph{Proc. Intl. Conf. on Pattern Recognit.
  \textup{(ICPR)}}, 2022, p. Under Review.

\bibitem{Raisi_2022_CRV}
Z.~Raisi and J.~Zelek, ``Occluded text detection and recognition in the wild,''
  in \emph{Proc. Conf. on Robots and Vision (CRV)}, 2022, p. In Press.

\bibitem{raisi_cvis_2022}
Z.~Raisi and J.~S. Zelek, ``End-to-end scene text spotting at character
  level,'' in \emph{Proc. Annual Conference on Vision and Intelligent Systems
  \textup{CVIS}}, 2021.

\bibitem{YOLO_2016}
J.~Redmon, S.~Divvala, R.~Girshick, and A.~Farhadi, ``You only look once:
  Unified, real-time object detection,'' in \emph{Proc. IEEE Conf. on Comp.
  Vision and Pattern Recognit.}, 2016, pp. 779--788.

\bibitem{fasterrcnn_ren2015}
S.~Ren, K.~He, R.~Girshick, and J.~Sun, ``Faster \textup{R-CNN}: Towards
  real-time object detection with region proposal networks,'' in \emph{Proc.
  Adv. in Neural Info. Process. Sys.}, 2015, pp. 91--99.

\bibitem{cut80_2014}
A.~Risnumawan, P.~Shivakumara, C.~S. Chan, and C.~L. Tan, ``A robust arbitrary
  text detection system for natural scene images,'' \emph{Expert Syst. with
  Appl.}, vol.~41, no.~18, pp. 8027--8048, 2014.

\bibitem{RNN_1986}
D.~E. Rumelhart, G.~E. Hinton, and R.~J. Williams, ``Learning representations
  by back-propagating errors,'' \emph{nature}, vol. 323, no. 6088, pp.
  533--536, 1986.

\bibitem{shahab2011icdar}
A.~Shahab, F.~Shafait, and A.~Dengel, ``\textup{ICDAR} 2011 robust reading
  competition challenge 2: Reading text in scene images,'' in \emph{Proc. Int.
  Conf. on Doc. Anal. and Recognit.}, 2011, pp. 1491--1496.

\bibitem{shi2018aster}
B.~Shi, M.~Yang, X.~Wang, P.~Lyu, C.~Yao, and X.~Bai, ``Aster: An attentional
  scene text recognizer with flexible rectification,'' \emph{IEEE Trans.
  Pattern Anal. Mach. Intell.}, 2018.

\bibitem{icdar2019}
Y.~Sun, Z.~Ni, C.-K. Chng, Y.~Liu, C.~Luo, C.~C. Ng, J.~Han, E.~Ding, J.~Liu,
  D.~Karatzas, \emph{et~al.}, ``\textup{ICDAR} 2019 competition on large-scale
  street view text with partial labeling--\textup{RRC-LSVT},'' \emph{arXiv
  preprint arXiv:1909.07741}, 2019.

\bibitem{transformer_survey_2020}
Y.~Tay, M.~Dehghani, D.~Bahri, and D.~Metzler, ``Efficient transformers: A
  survey,'' \emph{arXiv preprint arXiv:2009.06732}, 2020.

\bibitem{attention_vaswani2017}
A.~Vaswani, N.~Shazeer, N.~Parmar, J.~Uszkoreit, L.~Jones, A.~N. Gomez,
  {\L}.~Kaiser, and I.~Polosukhin, ``Attention is all you need,'' in
  \emph{Advances in neural information processing systems}, 2017, pp.
  5998--6008.

\bibitem{wang2010word}
K.~Wang and S.~Belongie, ``Word spotting in the wild,'' in \emph{Proc. Eur.
  Conf. on Comp. Vision}.\hskip 1em plus 0.5em minus 0.4em\relax Springer,
  2010, pp. 591--604.

\bibitem{yao2012detecting}
C.~Yao, X.~Bai, W.~Liu, Y.~Ma, and Z.~Tu, ``Detecting texts of arbitrary
  orientations in natural images,'' in \emph{Proc. IEEE Conf. on Comp. Vision
  and Pattern Recognit.}, 2012, pp. 1083--1090.

\bibitem{CTW_1500_yuliang2017}
L.~Yuliang, J.~Lianwen, Z.~Shuaitao, and Z.~Sheng, ``Detecting curve text in
  the wild: New dataset and new solution,'' in \emph{arXiv preprint
  arXiv:1712.02170}, 2017.

\bibitem{TESTR_2022}
X.~Zhang, Y.~Su, S.~Tripathi, and Z.~Tu, ``Text spotting transformers,''
  \emph{arXiv preprint arXiv:2204.01918}, 2022.

\bibitem{vpr4}
X.~Zhang, L.~Wang, and Y.~Su, ``Visual place recognition: A survey from deep
  learning perspective,'' \emph{Pattern Recognition}, vol. 113, p. 107760,
  2021.

\bibitem{deformableDETR_2020}
X.~Zhu, W.~Su, L.~Lu, B.~Li, X.~Wang, and J.~Dai, ``Deformable detr: Deformable
  transformers for end-to-end object detection,'' \emph{arXiv preprint
  arXiv:2010.04159}, 2020.

\end{thebibliography}
}


\section{Appendix}
\label{sec:appendix}

The TextPlace \cite{textplace_2019} model uses the pre-trained model of Textboxes++ \cite{liao2018textboxes++} algorithm as the main text extraction in their framework for the VPR application. In this section, we conduct additional experiments to compare our proposed model with Textboxes++ \cite{liao2018textboxes++}; To that effect, we provide quantitative and qualitative results to show how our model performs for text instances that appear in the wild images using the benchmark dataset, ICDAR15 \cite{karatzas2015icdar}, as in \cite{liao2018textboxes++}. The ICDAR15 is a challenging dataset that contains various indoor and outdoor multi-oriented text instances. Like most of the images in the VPR applications, this dataset has a wide variety of blurry and low-resolution text. 
The text instances in this dataset are annotated in quadrilateral bounding box annotations.

\paragraph{Quantitative Comparison.}
Table \ref{tab:textcompar} shows the quantitative comparison of our model and Textboxes++ using  the well-known text detection and end-to-end text spotting evaluation metrics \cite{karatzas2015icdar}. 
As seen, our proposed approach outperformed the \cite{liao2018textboxes++} in both detection and end-to-end spotting tasks. It achieves an H-mean detection performance of $86.5\%$ compare to $82.9\%$ in \cite{liao2018textboxes++}. It also surpasses the Textboxes++ method with a large margin of $\sim 16\%$ in end-to-end F-measure performance. 
Furthermore, as described in \S \ref{sec:experiment}, our model is more suitable for real-time detection and recognition as it provides a better FPS.
These performances confirm our models' well generalization and efficiency on challenging and unseen VPR dataset, SCTP (see Table \ref{tab:quant}).

\begin{figure*}[t]
    \centering
    \includegraphics[width=\textwidth ]{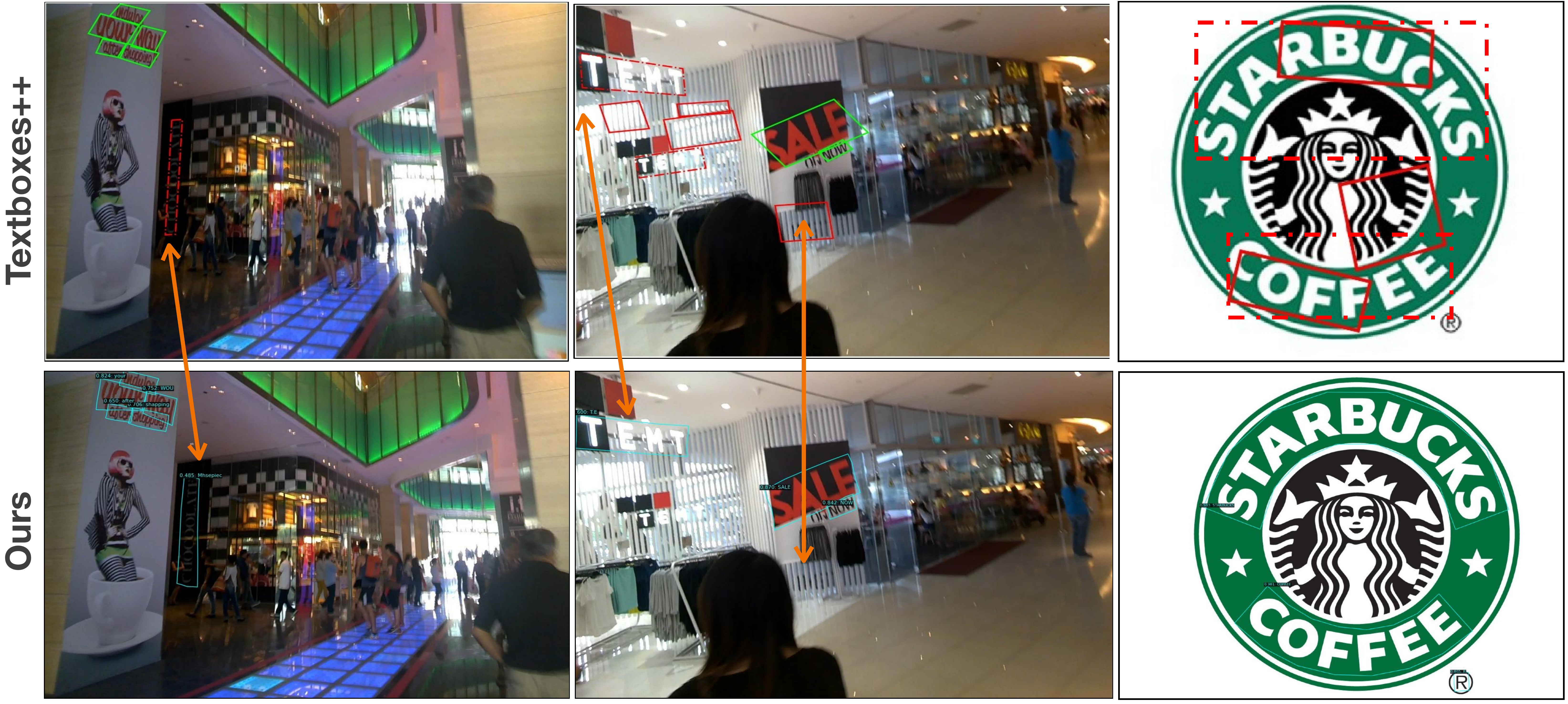}
    \caption{Comparison between our proposed end-to-end model (bottom row) with Textboxes++ \cite{liao2018textboxes++} algorithm (top row). The red boxes in the top row images show the failure cases (Images are taken from \cite{liao2018textboxes++}), and the cyan text and boxes show our results. The orange arrows point to text regions where our model could successfully predict failure text instances of Textboxes++.}
    \label{fig:qualcomp}
\end{figure*}

\paragraph{Qualitative Comparison.}
To see how our proposed algorithm performs in challenging cases of the ICDAR15 dataset, we also provide a qualitative comparison of our model with failure cases in \cite{liao2018textboxes++}. As shown in Figure \ref{fig:qualcomp}, Our method successfully predicted most of the failure cases.
Since text instances in the wild images usually appear with arbitrary shapes, this is important to use a model that better captures any shape of the scene text. The results in the last column in Figure \ref{fig:qualcomp} also show that our proposed pipeline is capable of accurately outputting polygon bounding boxes for curved text instances, whereas Textboxes++ fails to detect.

\begin{table}[h]
\vspace{-5pt}
    \centering
    \caption{Quantitative comparison of our model with Textboxex++ \cite{liao2018textboxes++} using ICDAR15 \cite{karatzas2015icdar} dataset. P, R, H, and F mean Precision, Recall, H-mean, and F-measure, respectively. E2E denotes end-to-end text spotting, and FPS is Frames per second. Best performance is highlighted in \textbf{bold}.}
\resizebox{\linewidth}{!}{%
\begin{tabular}{l|ccc|c|c}
\hline
\multirow{2}{*}{Model} & \multicolumn{3}{c|}{Detection}                 & E2E &\multirow{2}{*}{FPS}          \\ \cline{2-5}
                       & P             & R             & H             & F             \\ \hline
TextBoxes++ \cite{liao2018textboxes++}            & 87.8          & 78.5          & 82.9          & 51.9 &2.3         \\
\baseline{\textbf{Ours}}                   & \baseline{\textbf{90.2}} & \baseline{\textbf{83.1}} & \baseline{\textbf{86.5}} & \baseline{\textbf{68.2}}&\baseline{\textbf{11.0}} \\ \hline
\end{tabular}
}
    \label{tab:textcompar}
\end{table}

\end{document}